\documentclass[letterpaper, 10 pt, conference]{ieeeconf}  
\IEEEoverridecommandlockouts                              

\title{\LARGE \bf
Design and Evaluation of a Novel Cable-Driven Gripper with Perception Capabilities for Strawberry Picking Robots
}

\usepackage{amsmath}
\usepackage{amssymb}
\usepackage{graphicx}
\usepackage{subfig}

\usepackage{color}
\newcommand{\vtxt}[1]{\textcolor{red}{ \textit{{}}}} 

\author{Ya Xiong$^{1}$, Pal Johan From$^{1}$ and Volkan Isler$^{2}$
\thanks{*This work was  supported by the Norwegian Centennial Chair}
\thanks{$^{1}$Ya Xiong and Pal Johan From are with Faculty of Science and Technology, Norwegian University of Life Sciences, {Aa}s, Norway  {\tt\small \{ya.xiong, pal.johan.from\}@nmbu.no}}%
\thanks{$^{2}$Volkan Isler is with Department of Computer Science, University of Minnesota, Minneapolis, Minnesota, USA {\tt\small isler@umn.edu}}}

\begin{document}

\maketitle
\thispagestyle{empty}
\pagestyle{empty}

\begin{abstract}

This paper presents a novel cable-driven gripper with perception capabilities for autonomous harvesting of strawberries. Experiments show that the gripper allows for more accurate and faster picking of strawberries compared to existing systems. The gripper consists of four functional parts for sensing, picking, transmission, and storing. It has six fingers that open to form a closed space to swallow a target strawberry and push other surrounding berries away from the target. Equipped with three IR sensors, the gripper controls a manipulator arm to correct for positional error, and can thus pick strawberries that are not exactly localized by the vision algorithm, improving the robustness. Experiments show that the gripper is gentle on the berries as it merely cuts the stem and there is no physical interaction with the berries during the cutting process. We show that the gripper has close-to-perfect successful picking rate when addressing isolated strawberries. By including internal perception, we get high positional error tolerance, and avoid using slow, high-level closed-loop control. Moreover, the gripper can store several berries, which reduces the overall travel distance for the manipulator, and decreases the time needed to pick a single strawberry substantially. The experiments show that the gripper design decreased picking execution time noticeably compared to results found in literature.
\end{abstract}

\section{INTRODUCTION}

Strawberry is a high-value fruit. According to market research conducted by IndexBox, the global strawberry marketed revenue amounted to 21,171 million USD in 2015 and it continues growing noticeably. However, strawberry production heavily relies on human labor, with high labor cost, especially in harvesting [1]. In Norway, picking cost takes up more than 40-60\% of the whole labor costs, depending on production method, type of strawberries and yield. To reduce production cost and increase production quality, several research groups are trying to use robots to decrease dependency on human labor in soft fruit production. However, harvesting strawberries efficiently and reliably has proven to be extremely hard for several reasons. First of all, strawberries are easily damaged and bruised, which requires gentle handling for picking [2]. Secondly, strawberry harvesting requires highly selective procedures [3], since the strawberries tend to ripen very unevenly, giving large variations in color and size, and requiring several passes for harvesting. Finally, strawberries tend to grow in clusters, which makes it hard to identify and pick individual strawberries [1].

During the last few years, many systems have been developed for autonomous harvesting of soft crops ranging from cucumber [4] and tomato harvesting robots [6] to sweet pepper [7], [16] and strawberry picking apparatuses [1], [3]. Generally, strawberry picking robots should combine four subsystems: vision for detection, an arm for motion delivery, an end effector for picking, and finally a platform to increase the workspace to the size of the farm. The end effector is a critical component in this robotic system as it should allow for suitable manipulation [5] and pick the strawberries in a gentle and efficient way. An appropriate gripper design can enhance system operation stability and efficiency substantially [6]. Thus far, researchers have developed several types of end effectors for strawberry picking, such as scissor-like cutters [3], cutters with suction device [8], as well as force-limit grasping grippers [2]. As the position of the strawberry stem (picking point) is difficult to detect [9], [10], especially in a cluster, the scissor-like end effectors require a relatively hard vision problem to be solved. It is also easy to cut more than one stem at the time and unintentionally pick green strawberries. Force controlled grippers are also hard to use as one would very easily bruise fragile strawberries [3]. 

In this paper, we present a cable driven non-touch picking gripper with a storage container. The main advantage of the design is that the gripper is able to pick strawberries in a simple cluster by cutting the stem without needing to know the exact stem position. The strawberries are therefore picked without touching the actual strawberry, which is beneficial for the quality of the berries.

\section{DESIGN}
\begin{figure*}[!tbp]
  \centering
  \subfloat[Inner view of picking and transmission mechanism.]{\includegraphics[scale=0.52]{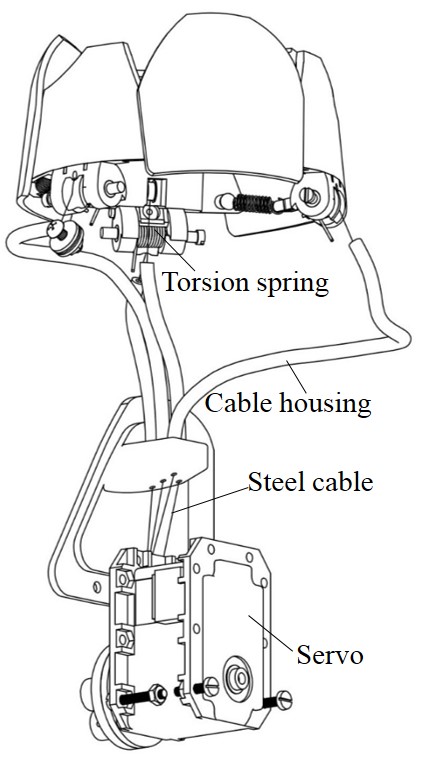}\label{fig:f2}}
  \subfloat[Perspective view of the gripper.]{\includegraphics[scale=0.51]{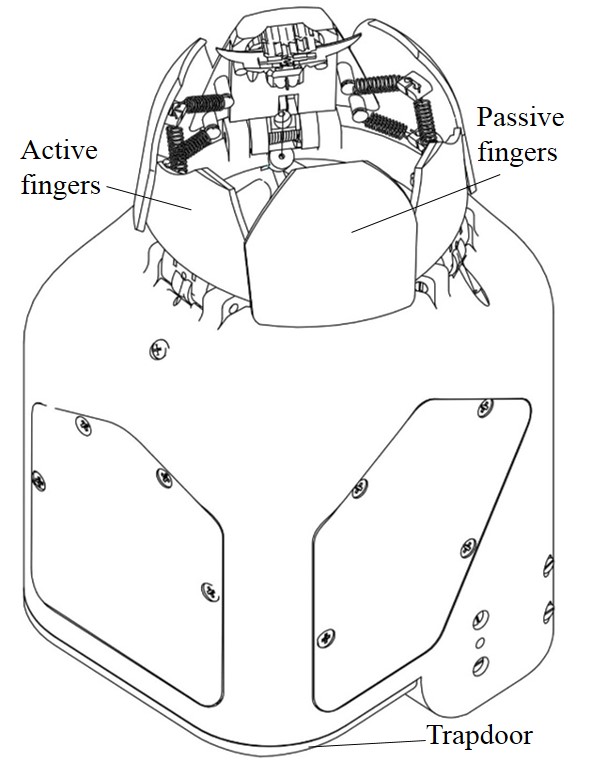}\label{fig:f1}}
  \subfloat[Inside bottom view of prototype.]{\includegraphics[scale=0.51]{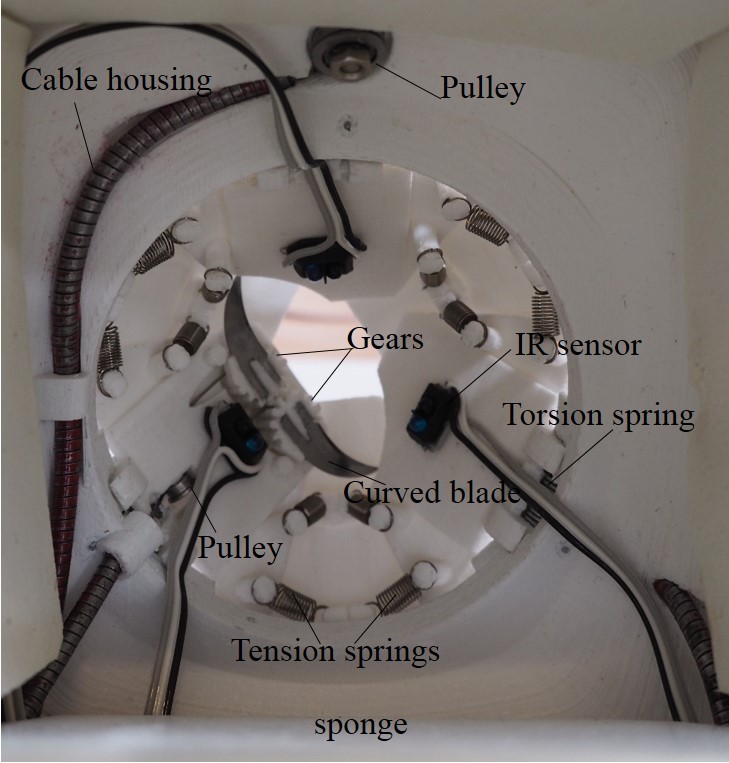}\label{fig:f3}}
  \caption{The cable driven gripper.}
\end{figure*}
\begin{figure}[thpb]
    \centering
    \includegraphics[scale=0.6]{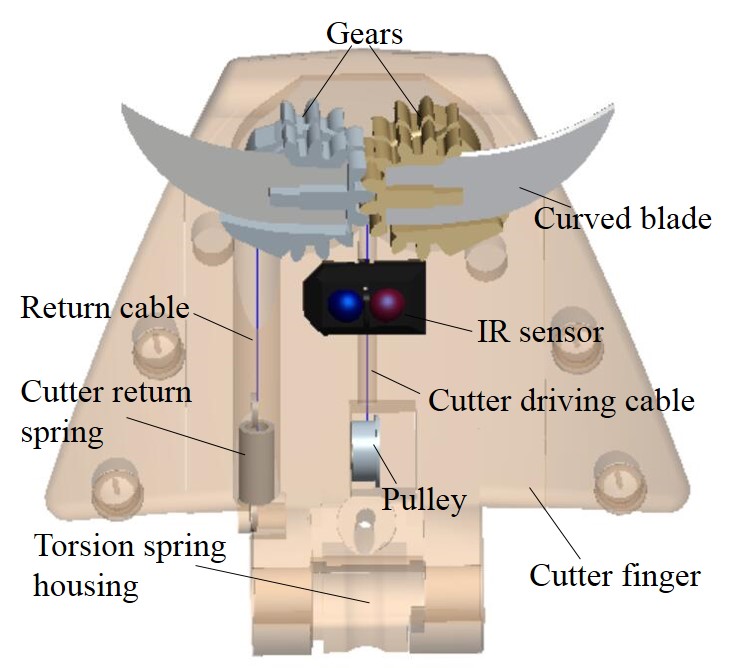}
    \caption{3D model of cutter mechanism}
    \label{figurelabel}
\end{figure}
The main objectives and challenges of the gripper design were i) gentle or no interaction with fragile strawberries, ii) separate strawberries that are in clusters, iii) have a high tolerance for positional error, and iv) achieve high picking speeds. Gentle interaction means that the gripper should not damage the strawberry to be picked, nor other surrounding strawberries. Moreover, strawberries are growing in the 3D space, so if the robot is targeting a strawberry at the back, the scissor-like grippers presented in most of the literature will unintentionally cut front branches and possibly green strawberries. Therefore, we needed to developed a gripper that can pick from below and thus only ``swallow" the targeted strawberry. 

The proposed gripper is illustrated in Fig. 1. The overall size of the gripper including the storage container is $110\times130\times178mm$. It can be divided into four parts by its function:  picking, transmission, sensing, and storing. The picking mechanism consists of three active fingers, three passive cover fingers and a cutter mechanism. Inspired by skilled workers who use fingers to gently pull out a strawberry and nails to cut the stem, the gripper opens the fingers to swallow a strawberry, then closes the fingers, and finally the cutter inside of the fingers rotates quickly to cut the stem. Thus, the cutter is hidden inside of the fingers to avoid damage to the target strawberry as well as the surrounding ones. While closing the fingers, the fingers can push the stem to the cutting area. If in a cluster, the fingers can open based on the strawberry size and push other surrounding strawberries away so that only one strawberry is swallowed into the container. To realize this, it is necessary to design a mechanism that allows for several parts to open or close simultaneously that form a closed ring to enclose the strawberry.  Preferably, these moving parts could be controlled by only one motor to reduce cost and space. Different from the complex classic Iris mechanism [11], we used three active driving fingers and three passive driving fingers (cover fingers) to swallow the target strawberry. Four small tension springs were used to keep the cover fingers adhering to the fingers regardless of their rotation. 

For transmission, it is important to keep the space around fingers and cutters as small as possible. Gear transmission can provide an outstanding result for simultaneous motion. However, this kind of transmission needs much space below the fingers and it would be too complex to control the three fingers using only one motor. Therefore, similar to some advanced robotic hands and snake robots [12], [13], a cable-driven method was adopted since it is very well suited for remote transmission. Using this method, the gripper could have more space under the fingers for picking and storing strawberries and the motor can be placed relatively far away from the joints. Extra flexible steel cables were selected so the direction can be changed easily using normal bicycle cable housings and pulleys, as seen in Fig. 1(c). To reduce cost and complexity, only one servo with different rotation directions that control both the fingers and cutter was used , as they never work simultaneously. The fingers open by driven cables and close by torsion springs. As shown in Fig. 2, the cutter, comprised of two curved blades, was mounted on a pair of small gears. One active driving gear was pulled by a cable so the cutter can close and the other passive driving gear, connected to a return tension spring, was used to open the cutter. As a consequence, at the central position, the servo turned anticlockwise to open fingers and clockwise to rotate cutter. Slackness of steel cable might happen on one mechanism when the other side of the mechanism was actuated, which may result in the steel cables running off the pulley. This could be solved by adding additional tension mechanisms, but this would increase device complexity and space. Therefore, cable ends were mounted on a cylindrical motor connector rather than a pulley.  Also, only two pulleys were used, one for the far-located finger and the other one for the cutter mechanism, but the steel cables on these pulleys were limited by the outer structure to avoid running off. 

One advantage of this gripper design is that it has a high positional error tolerance, so it does not need the exact stem position. As discussed above, the gripper is able to use the fingers to push the stem into the cutting area, eliminating errors in the localization of the strawberry and other uncertainties. This mechanism is efficient, but at the same time we experienced that the cutting is a lot more robust if the strawberry is located in the middle of the container. Therefore, the gripper is equipped with three internal Vishay TCRT5000 infrared (IR) sensors for active optimal cutting position control and also to control the length of the stem. The control algorithm will be explained in Section IV. 

When a strawberry has been picked, a container below the fingers is used to collect and store the strawberries. With this container, the robot can pick strawberries without moving the arm to store the strawberry for every one picked, but rather store them in the container. Once the container is full, a trapdoor on the bottom would be opened to dispense strawberries into a packing box. A second servo motor is used to control the trapdoor. A tension spring is applied to keep the lid closed when servo power was turned off. The container can store 7-12 strawberries depending on the size. To minimize the damage to strawberries, the container has an inclined dropping board to break the fall of strawberries after they have been cut. Most importantly, on the inside of the container, pieces of soft sponge was used to protect strawberries. Except for the servos, sensors and transmission system, all other parts of the gripper prototype were 3D printed.

\section{KINEMATIC ANALYSIS}
\begin{figure}[thpb]
    \centering
    \includegraphics[scale=0.65]{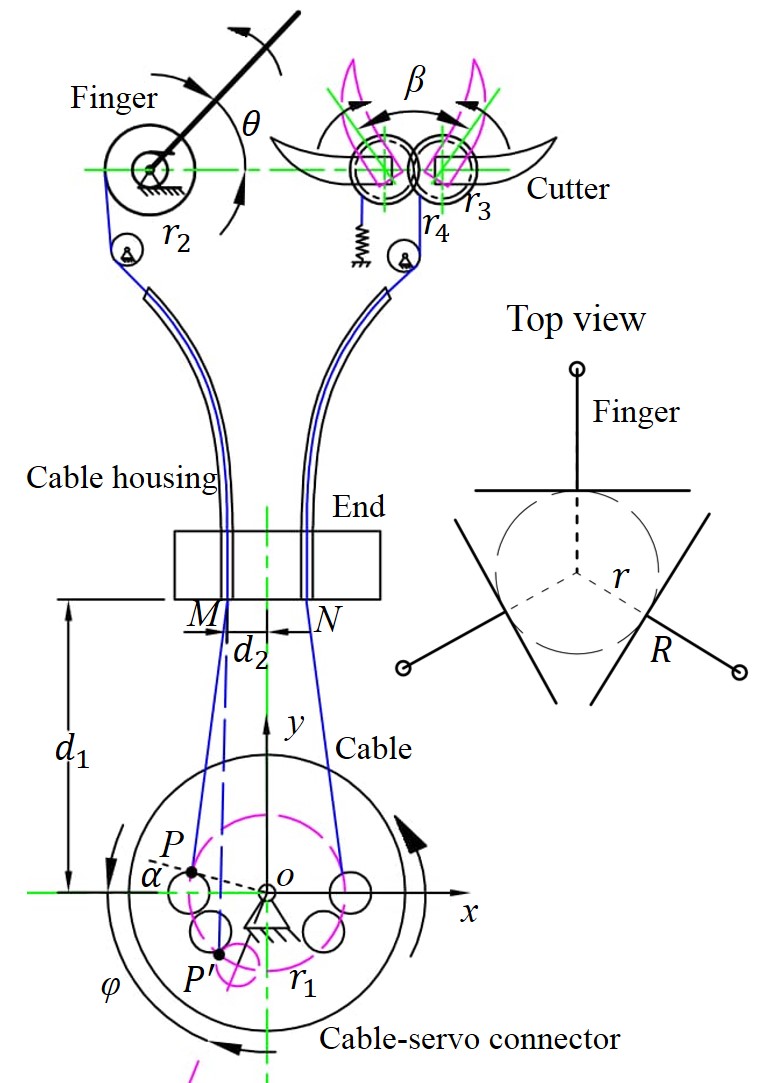}
    \caption{Kinematic schematic of gripper transmission}
    \label{figurelabel}
\end{figure}
The purpose of kinematic analysis is to calculate the position of the fingers and cutter angle from the servo angular displacement. With this, once the strawberry shoulder diameter (maximum) is determined (either by vision system or gripper sensor scanning), the servo can rotate an angle calculated from the size of the strawberry to swallow the target with an appropriate opening.  Fig. 3 shows a schematic drawing of the kinematics of the gripper transmission. The key point is to identify how cable extension and retraction affect both fingers, cutter and servo. The length of the cable on the servo side ($P'M$) can be expressed as:
\begin{equation}  
\left\{  
             \begin{array}{lr}  
             P'(x)=r_{1}\cos(\varphi-\alpha) &  \\  
            P'(y)=r_{1}\sin(\varphi-\alpha) &    
             \end{array}  
\right.  
\end{equation}  
$$
D=l_{P'M}=\sqrt{(P'(x)-(-d_{2}))^2+(P'(y)-d_{1})^2} \eqno{(2)},
$$
where, $\varphi$ is the angular displacement of the servo, while $r_{1}$ and $\alpha$ are the rotation radius and initial angle of $oP$ to $x$-axis the of connected point $P'$. $d_{1}$ and $d_{2}$ are the offsets of point $M$ to origin $o$. Then the angle of finger $\theta$ is obtained as:
$$
 \theta=\dfrac{D-l_{PM}}{r_{2}}+\theta_{0} \eqno{(3)},
$$
where $l_{PM}$ and $\theta_{0}$ are the initial length and angle of $P'M$ and $\theta$, respectively. $r_{2}$ denotes the radius of finger pulley. With $\theta$, from the space `Top view', we can obtain the radius ($r$) of fingers open size:
$$
 r=R-l_{fin}\cos(\theta) \eqno{(4)},
$$
where $R$ is the radius of finger joints circle, and $l_{fin}$ is the length of the finger. As shown in Fig. 4, the red line indicates $r$ to $\varphi$. Within the limit open size $r\in [0,40mm]$,  $\varphi\in [0,28.3^\circ]$ accordingly. Similarly, cutter blades angle $\beta$ to $\varphi$ is illustrated as the blue line. In addition, since the blue line is a straight line and the red line is almost straight, both the angular velocity of cutter and fingers open size velocity have a relative fixed relationship with servo angular velocity, which is easy for speed control.  

\begin{figure}[thpb]
    \centering
    \includegraphics[scale=0.47]{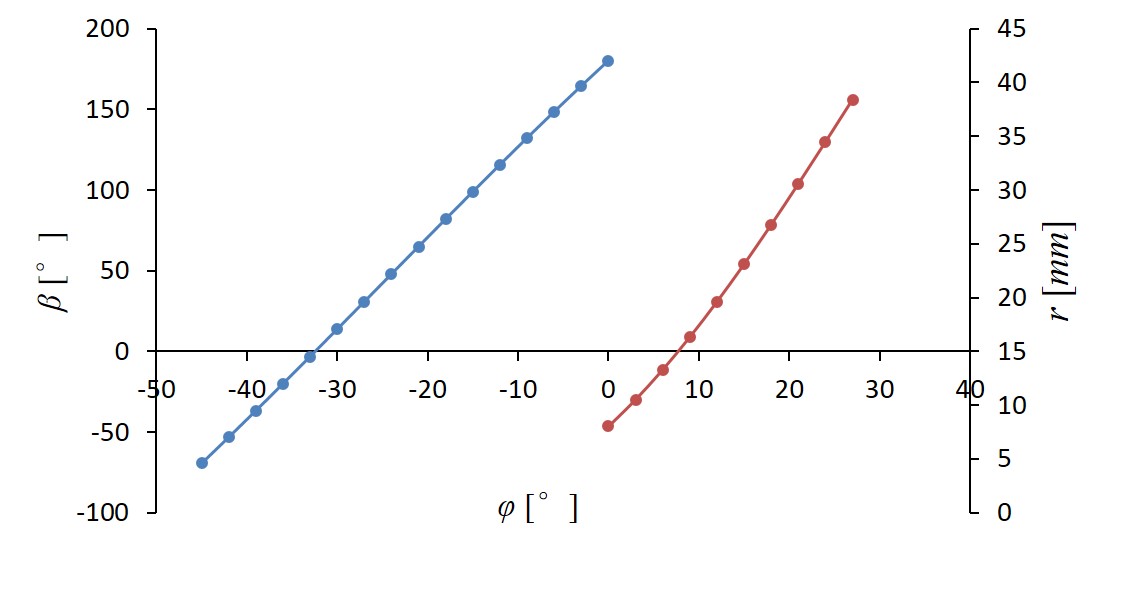}
    \caption{Curve of kinematic analysis result: when servo angular displacement $\varphi>0$, the red line indicates how the fingers open size $r$ changes along $\varphi$ (almost linear), while $\varphi<0$, the blude line indicates the cutter blades angle $\beta$ has a linear relationship with $\varphi$. }
    \label{figurelabel}
\end{figure}
\section{OPTIMAL CUTTING POSITION AND STRAWBERRY STEM LENGTH CONTROL}

\subsection{Optimal Cutting Position Control}
To achieve active cutting position, it is essential to know the position of strawberry with respect to the gripper. To realize this, the first thing is to obtain the distance from the strawberry to the IR sensor (defined as $mdp$). However, IR sensors are easily disturbed by sunlight, as shown in Fig. 5(a).  Thus, a simple algorithm has been used, turning on IR LED for 0.5 $ms$ to get a raw distance and then turning off for the same period to get the noise. Subtracting noise from the raw distance gives the filtered datum. After filtering, the gray line had a stable response regardless of light change. Then a calibration test has been conducted to convert IR output analog value into distance by using strawberry as obstacle, as seen in Fig. 5(b).  
\begin{figure}[!tbp]
  \centering
  \subfloat[Signal filtering test.]{\includegraphics[scale=0.28]{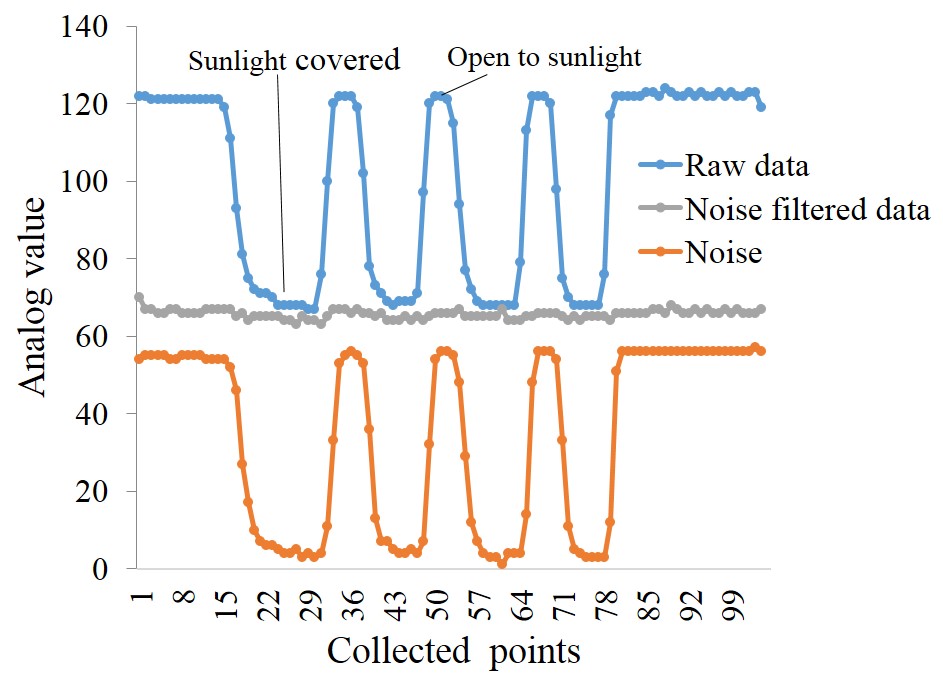}\label{fig:f5}}
  \hfill
  \subfloat[Calibration by using strawberry as obstacle.]{\includegraphics[scale=0.28]{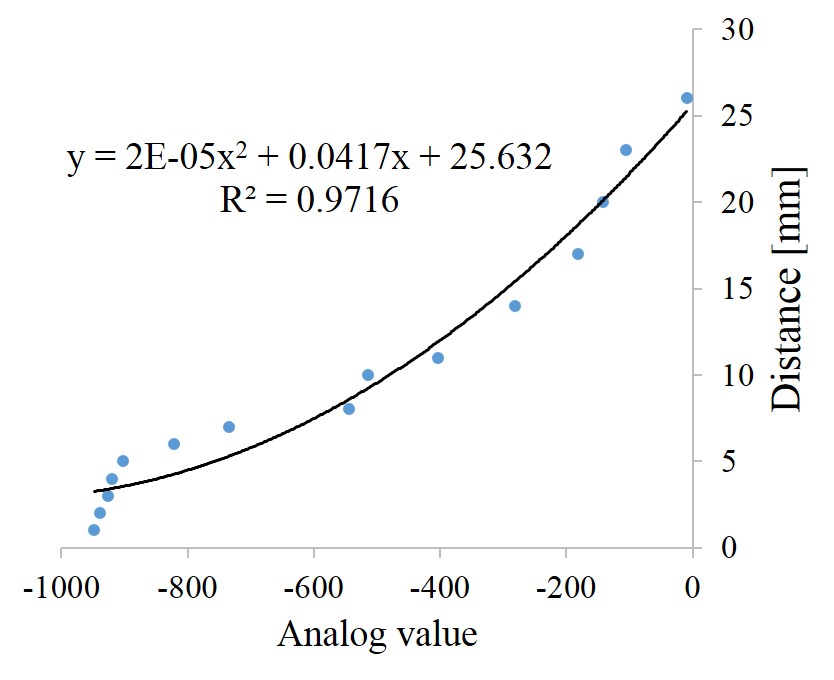}\label{fig:f6}}
  \caption{Curve of IR sensor signal filtering and calibration.}
\end{figure}

\begin{figure}[!tbp]
  \centering
  \subfloat[Front view sketch.]{\includegraphics[scale=0.42]{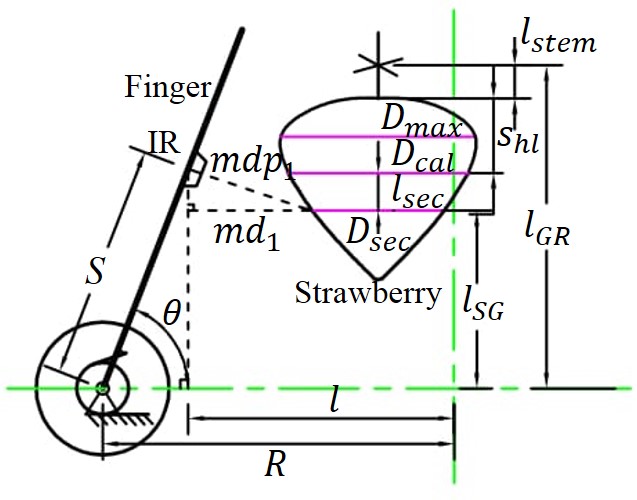}\label{fig:f5}}
  \hfill
  \subfloat[Section view sketch.]{\includegraphics[scale=0.42]{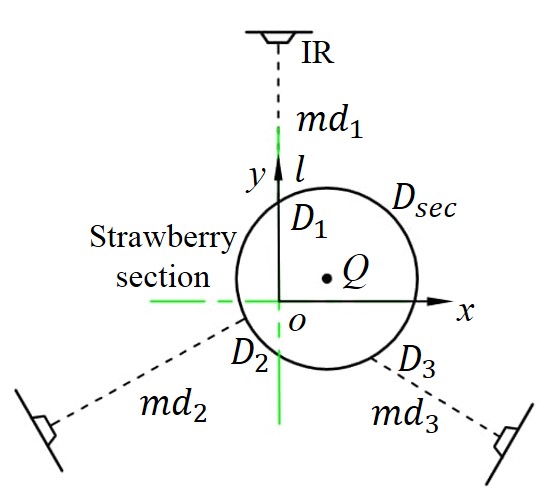}\label{fig:f6}}
  \caption{Schematic of strawberry optimal cutting position control.}
\end{figure}

Fig. 6 shows strawberry position in the gripper by both front view and section view. Finger rotation angle $\theta$ has already been obtained in section III, so projection distance $md$ can be described as: 
\begin{equation} 
             \begin{array}{lr}  \tag{5}
             md_{1}=mdp_{1}\sin(\theta) &  \\  
            md_{2}=mdp_{2}\sin(\theta)& \\
            md_{3}=mdp_{3}\sin(\theta).&   
             \end{array}  
\end{equation}

Also, the distance ($l$) between IR center and gripper center is:  
$$
l=R-S\cos(\theta) \eqno{(6)}, 
$$
Where, $S$ is the mounting position of IR center to finger joint. Then, as shown in Fig. 6(b), coordinates of detected points $D_{1},D_{2},D_{3}$ in $oxy$ can be expressed as:
\begin{equation}   
             \begin{array}{lr}  \tag{7}
             D_{1}(0, l-md_{1}) &  \\  
            D_{2}(-\cos(\dfrac{\pi}{6})(l-md_{2}), -\sin(\dfrac{\pi}{6})(l-md_{2}))& \\
            D_{3}(\cos(\dfrac{\pi}{6})(l-md_{3}), -\sin(\dfrac{\pi}{6})(l-md_{3}))&   
             \end{array}  
\end{equation}

Based on the above three points, assuming strawberry section is a strict circle, the centroid ($Q$) and diameter ($D_{sec}$) can be obtained as:
\begin{equation}   
             \begin{array}{lr}  \tag{8}
             a=2(D_{2}(x)-D_{1}(x)) &  \\  
            b=2(D_{2}(y)-D_{1}(y))& \\
            c=D_{2}^2(x)+D_{2}^2(y)-D_{1}^2(x)-D_{1}^2(y)& \\
           d=2(D_{3}(x)-D_{2}(x))& \\
           e=2(D_{3}(y)-D_{2}(y))& \\
           f=D_{3}^2(x)+D_{3}^2(y)-D_{2}^2(x)-D_{2}^2(y)& 
             \end{array}   
\end{equation}
\begin{equation} 
\Rightarrow \left\{ 
             \begin{array}{lr}  \tag{9}
            offset_{x}= Q_{x}=\dfrac{bf-ec}{bd-ea} &  \\  
           offset_{y}= Q_{y}=\dfrac{dc-af}{bd-ea}& \\
            D_{sec}=2\sqrt{(Q_{x}-D_{1}(x))^2+(Q_{y}-D_{1}(y))^2}&   
             \end{array}  
\right.  
\end{equation}

The distance data were collected every 50 $ms$, which is quick enough for continuous measuring and closed loop control. We tested the measurement ability by moving strawberries up and down continuously and recorded the maximum diameter (shoulder). The comparative measurements result is shown in Fig. 7. The result indicated the gripper has a good measurement ability with average error of 0.81 $mm$, standard deviation 0.93 $mm$.
\begin{figure}[thpb]
    \centering
    \includegraphics[scale=0.5]{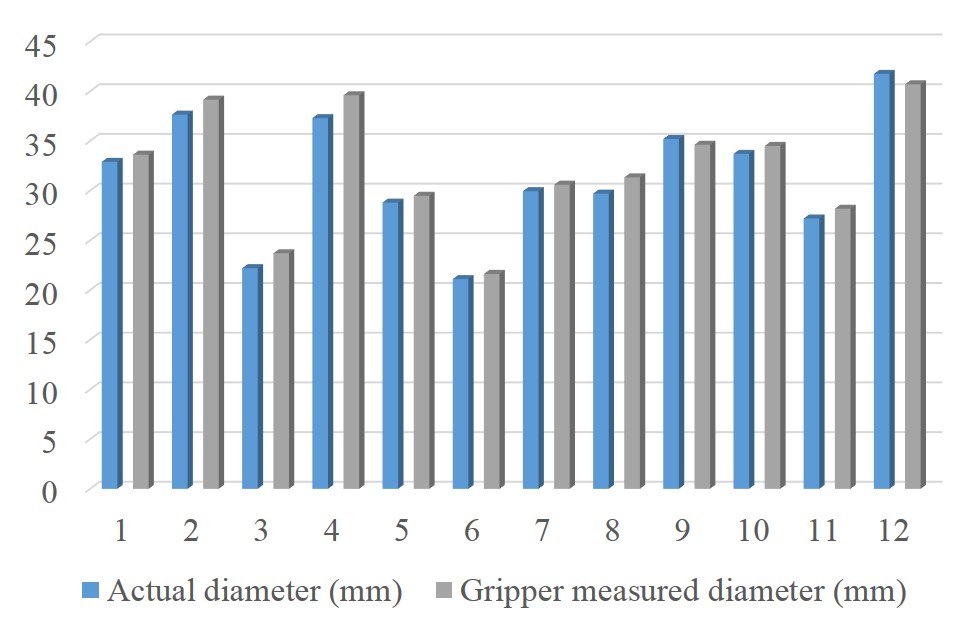}
    \caption{Gripper measurement ability test}
    \label{figurelabel}
\end{figure}
\begin{figure}[thpb]
    \centering
    \includegraphics[scale=0.45]{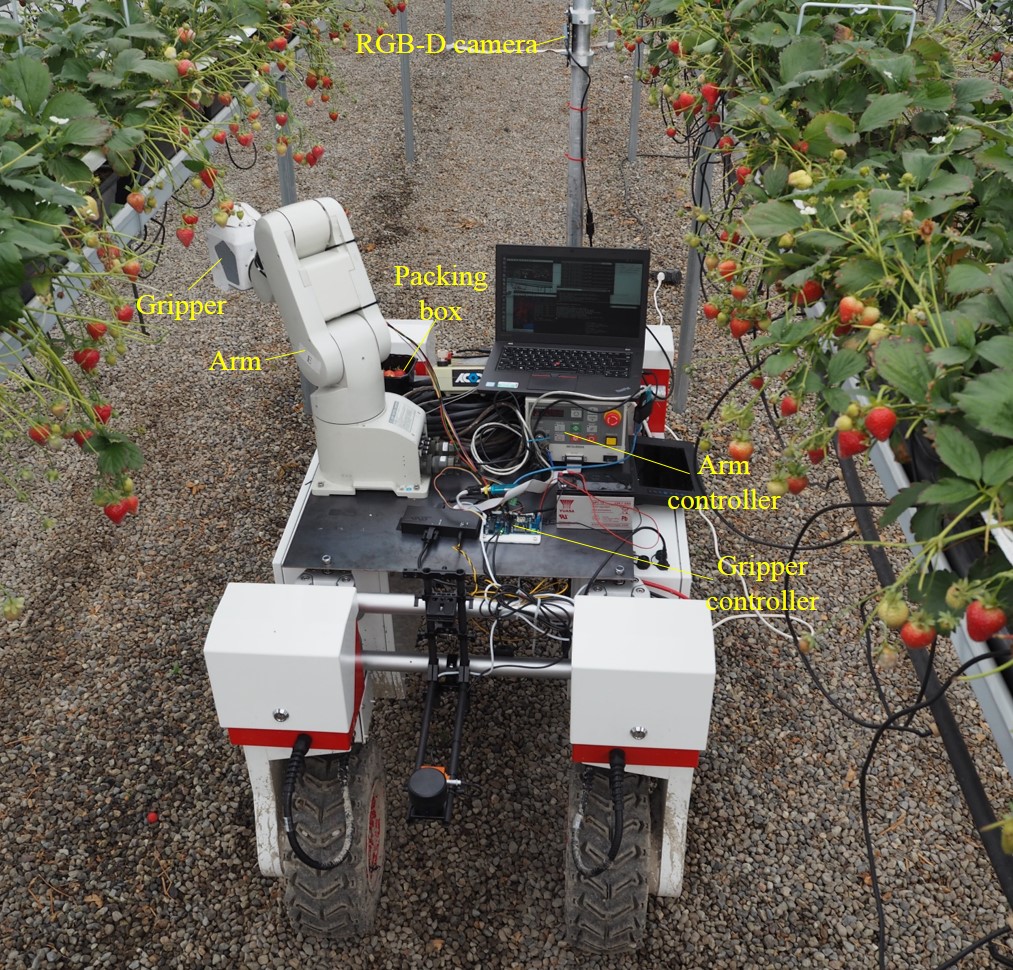}
    \caption{Field test setup}
    \label{figurelabel}
\end{figure}

Fig. 8 shows a testing setup of strawberry picking that has been built in strawberry tunnels. In addition to the gripper, the system mainly consists of a Thorvald platform  [14], [15], Intel RGB-D camera, and Mitsubishi RV-2AJ arm. To control the strawberry position with respect to the gripper at a target position ($target_{x}$, $target_{y}$), the errors $error_{x}$ and $error_{y}$ in equation (10) were converted into arm frame based on geometry dimensions. Then two parallel PID loops were used for $error_{x}$ and $error_{y}$, respectively.
\begin{equation} 
 \left\{ 
             \begin{array}{lr}  \tag{10}
            error_{x}= offset_{x}+target_{x}  &  \\  
           error_{y}= offset_{y}+target_{y} & 
             \end{array}  
\right.  
\end{equation}

The control results are shown in Fig. 9 and Fig. 10. The arm could move the gripper gently to place the strawberry at the target cutting position. With initial error 6.34 $mm$ and threshold error 0.1 $mm$ in Fig. 9, the relative stable settling time was around 3 $s$. In real picking, it is not necessary to be that  accurate as the gripper itself has a high mechanical tolerance. We used 1.5 $mm$ as error threshold and the average settling time was around 1 $s$. 
\begin{figure}[thpb]
    \centering
    \includegraphics[scale=0.5]{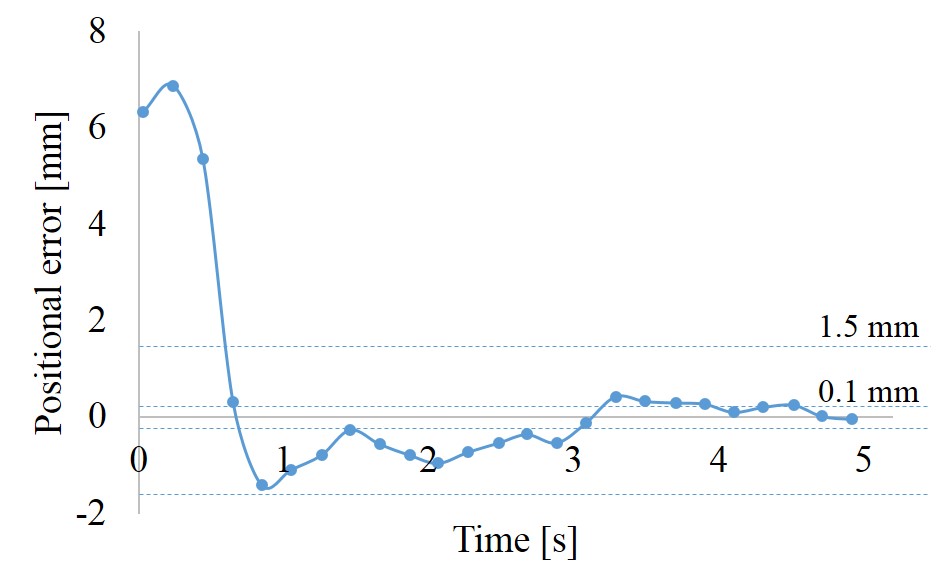}
    \caption{PID control test of strawberry position with respect to the gripper}
    \label{figurelabel}
\end{figure}
\begin{figure}[thpb]
    \centering
    \includegraphics[scale=0.61]{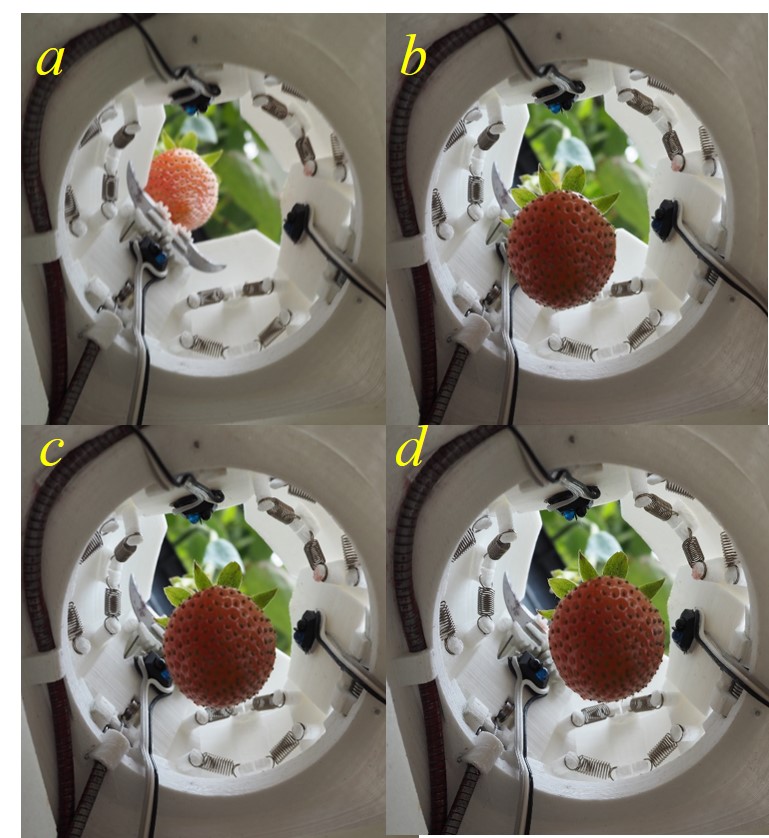}
    \caption{Real environment test of optimal cutting position control: from $a$ to $d$, the arm moved gripper gently to place strawberry at a target cutting position with respect to the gripper }
    \label{figurelabel}
\end{figure}

\subsection{Strawberry Stem Length Control}
In addition to $x$ and $y$ errors, it is also meaningful to control $z$ error. If this error is too big, the gripper will either cut the body of the strawberry or too long stem. As shown in Fig. 6(a), with the IR sensors, strawberry section diameter $D_{sec}$ and section height to joint $l_{SG}$ can be obtained. $l_{SG}$ is:
$$
l_{SG}=S\sin(theta)-Smdp_{1}\cos(\theta) \eqno{(11)}.
$$
Assuming strawberry shoulder diameter $D_{max}$ is determined, if strawberry section to its top distance $S_{hl}$ has a relationship with $D_{max}$, the top position will be identified. $D_{max}$ might be obtained by a camera from bottom view vision (Fig. 11(b)) or by measuring strawberry area and find the relationship between area and $D_{max}$ or by gripper IR sensor continuous scanning, which is another scope of research. Based on observation of strawberries in the tunnels (species: ``FAVORI"), most of the strawberries have similar shape properties, especially for the underpart that is almost triangle (Fig. 11(a)).  Hence, an investigation of strawberry shape properties has been implemented, as shown in Table I. The angles $\gamma$ of the underparts are very close, average value $52.72 ^\circ$. Also, the average ratios of  $D_{max}/D_{cal}$ and  $D_{max}/S_{hl}$ are 1.11 and 1.81, respectively.
\begin{figure}[!tbp]
  \centering
  \subfloat[Curve fitting of strawberry shape.]{\includegraphics[scale=0.47]{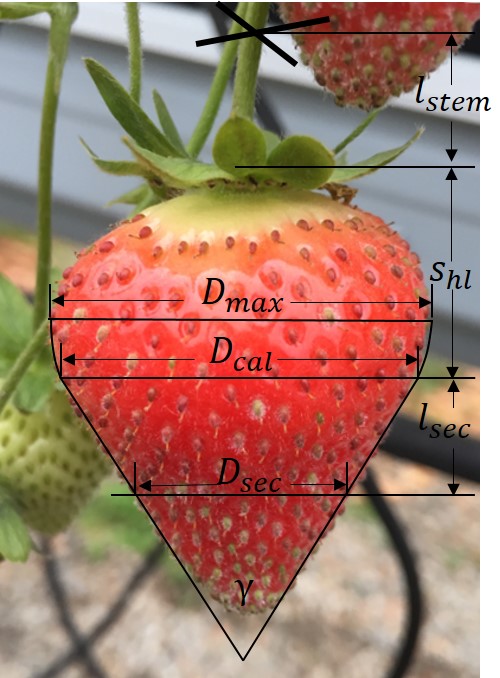}\label{fig:f5}}
  \hfill
  \subfloat[strawberry shoulder diameter detection from bottom view.]{\includegraphics[scale=0.47]{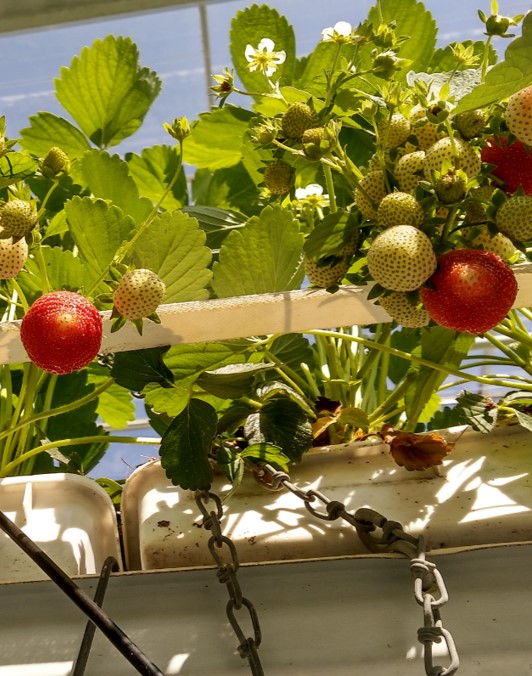}\label{fig:f6}}
  \caption{Strawberry shape properties investigation.}
\end{figure}

\begin{table}[h]
\centering
\caption{Strawberry shape properties investigation data}
\includegraphics[scale=0.58]{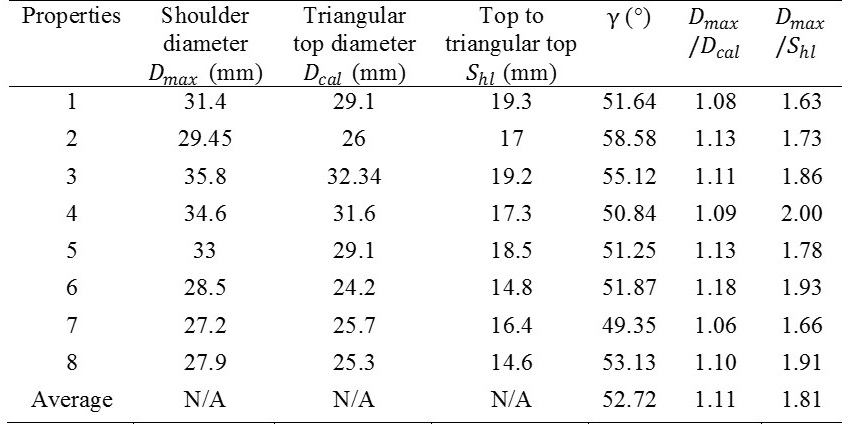}
\end{table}

With the above conditions, the length $l_{sec}$ from section $D_{sec}$ to triangle top section $D_{cal}$ can be obtained as:
\begin{equation} 
             \begin{array}{lr}  \tag{12}
             S_{hl}=D_{max}/1.81 &  \\  
            D_{cal}=D_{max}/1.11 &  \\
            l_{sec}=\dfrac{1}{2}\tan(\dfrac{\gamma}{2})(D_{cal}-D_{sec})& 
             \end{array}  
\end{equation}

Then offset to adjust current $z$ position $offset_{z}$ is:
$$
\Rightarrow offset_{z}=l_{stem}+S_{hl}+l_{sec}-(l_{GR}-l_{SG}) \eqno{(13)}
$$
where, $l_{GR}$ is a constant value of cutting position height to joint, while $l_{stem}$ is the cutting stem length that required. 

\section{EVALUATION EXPERIMENT}
Four sets of field experiments were performed to evaluate the performance of the proposed gripper. The first one was to examine the swallowing, separating and storage ability of the gripper. Fig. 13 shows a real environment example for picking 3 strawberries continuously. Strawberries were detected in Fig. 13(a); arm moved quickly to the bottom of the first strawberry (Fig. 13(b)); opened fingers and slow speed lifted up for searching strawberry and swallowed it (Fig. 13(c)); PID control to correct $error_{x}$ and $ error_{y}$ (Fig. 13(d)); corrected $offset_{z}$ to a required stem length (Fig. 13(e))); cut it gently in Fig. 13(f)); quickly moved to the second strawberry (Fig. 13(g)) and conducted the same operations; finally, the picked strawberries that has been stored in the container were placed gently to the packing box (Fig. 13(p)). During these procedures, it can be seen that the gripper was able to push other surrounded green strawberries and only swallowed the target. In the experiments, the average consumption time of continuous single strawberry picking was 7.49 $s$ (excluding detection, first and final arm travelling and placing strawberries), while including all procedures, the average time was 10.62 $s$ for picking one. As a consequence, the perception and storage ability enabled the robot to get out of visual servoing and travelling arm for single target, so the picking execution was much faster than the previous research strawberry picking 32.3 $s$ [3], sweet pepper picking 105.8 $s$ [16] and another sweet pepper research 35-40  $s$ [7], as well as tomato picking 23 $s$ [6]. The picking time for each strawberry will further decrease if more than three strawberries are picked before emptying the container. 

\begin{figure}[thpb]
    \centering
    \includegraphics[scale=0.47]{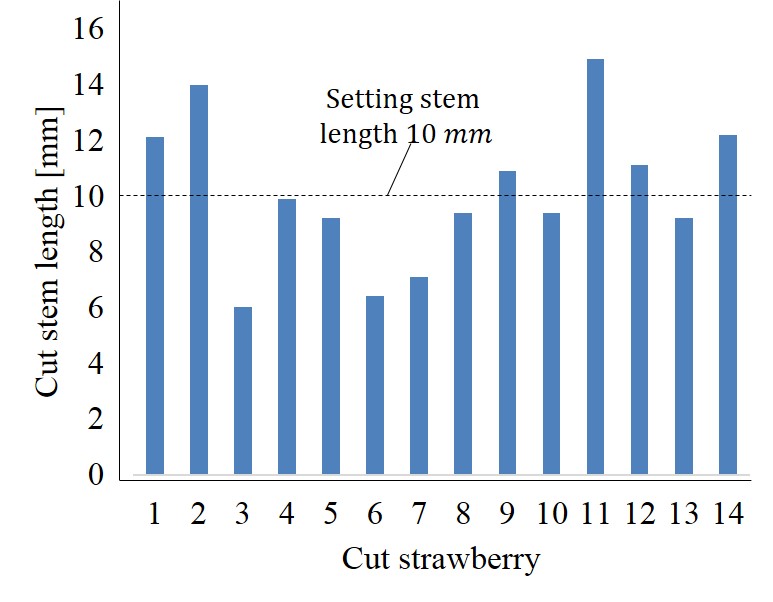}
    \caption{Stem length control test result.}
    \label{figurelabel}
\end{figure}
\begin{figure*}[!tbp]
    \centering
    \includegraphics[scale=0.47]{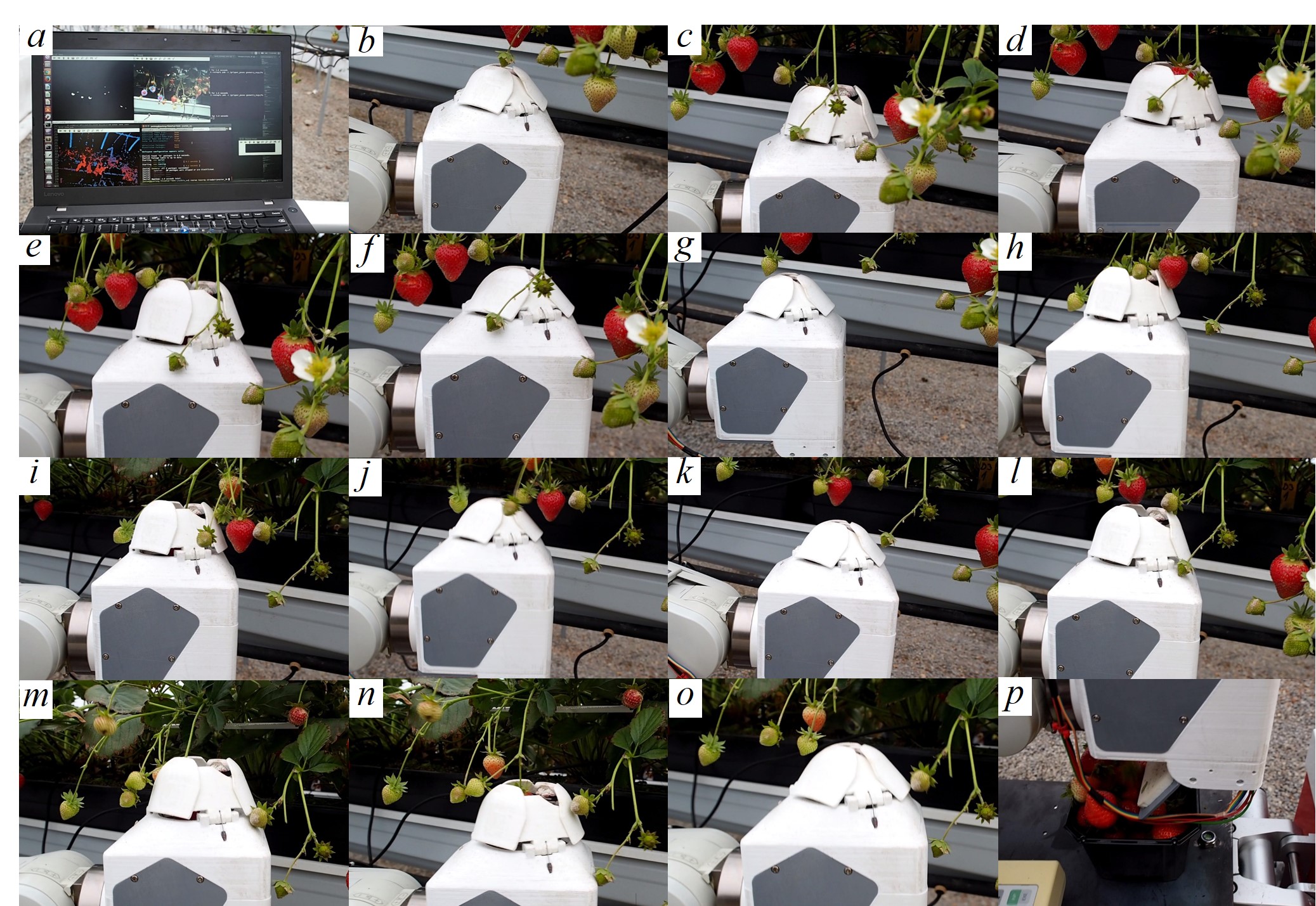}
    \caption{Picking procedures of continuous strawberry picking in field test }
    \label{figurelabel}
\end{figure*}

The second experiment was stem length control evaluation. Strawberry shoulder diameter $D_{max}$ was measured manually for stem length control test. As shown in Fig. (12), with a preset stem length of 10 $mm$, the actual average cut stem length was 10.13 $mm$, standard deviation 2.64 $mm$. The big errors were caused by inclined strawberries. Generally, if the gripper continuously scans two or more sections of the strawberry or employing more IR sensors, it is possible to obtain the orientation of the strawberry and even $D_{max}$, which is our future work.

The third test was conducted in simplified environment to identify the optimal blade cutting position and picking success rate for isolated strawberries. The stem diameter of cut strawberry varied from 1.7 $mm$ to 2.5 $mm$. The result is illustrated in Table II. By comparison, the optimal cutting position is the gripper origin, which is the central position of the cutter. The success rate for picking isolated strawberries was 96.77\% at the optimal position (previous strawberry picking research was around 70\% [3]), which demonstrated that the cutter was extremely robust for errors.
\begin{table}[h]
\centering
\caption{Picking isolated strawberries success rate with different cutting position}
\includegraphics[scale=0.62]{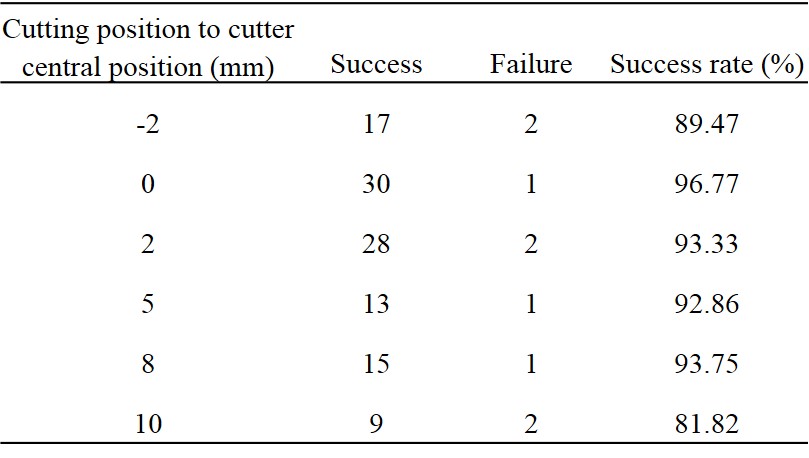}
\end{table}

Finally, a whole system picking test was performed in natural environment. Table III indicates the success rate with damage was 58.93\% while without damage was 53.57\%, which is higher than the similar research [16]. The failure was caused by many different aspects, including detection, hidden, arm reach region, dynamic disturbances and also the gripper. The main challenge for the gripper was when picking in dense clusters, the gripper “mouth” was easily to be covered by branches, leaves and green strawberries, so it was difficult to swallow the target strawberry. 
\begin{table}[h]
\centering
\caption{Whole system picking in unchanged natural environment}
\includegraphics[scale=0.52]{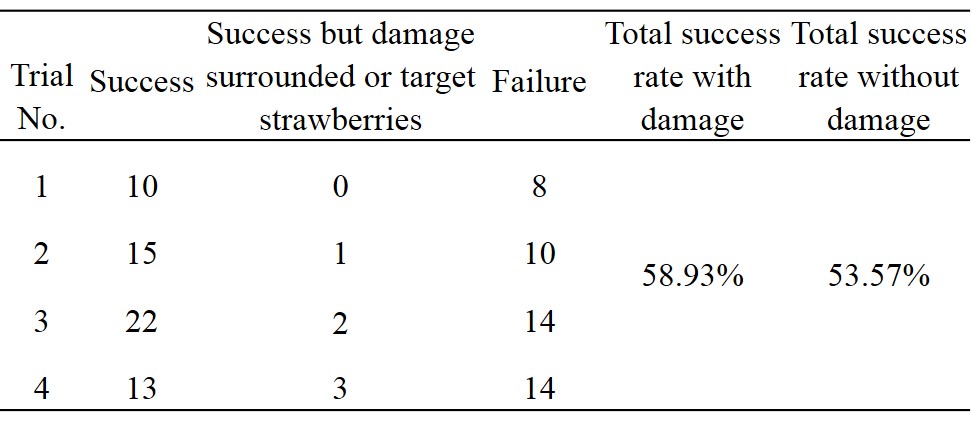}
\end{table}

\section{CONCLUSIONS}
We presented the design, analysis and early experimental results of a novel system for strawberry harvesting. One of the main novelties of the system is the gripper which is robust to uncertainties in the location and orientation of the strawberry. The gripper also has the capability to store several strawberries which reduces travel time for the manipulator arm. Experiments demonstrated that it has a very high success rate for picking isolated strawberries (96.77\%). Furthermore, with perception and storage abilities, in continuous picking, the gripper can pick a single strawberry in 7.49 s (the picking operation only). If we include all procedures, the average picking time is 10.62 s for picking one strawberry, which is much faster than previous research. An additional and important property of the gripper is that, based on the shape of the strawberry, the gripper can control the length of the stem that remains on the strawberry after cutting. 
 
For future work, we will develop the approach further to account for cases when the strawberries are located in dense clusters. At the moment, such cases are challenging for our system. 

\section*{ACKNOWLEDGMENT}

We thank Mr. Simen Myhrene from the Myhrene farm for providing the strawberry tunnels and accommodation to conduct field experiments.


\begin{thebibliography}{99}

\bibitem{c1} Yamamoto, S., Hayashi, S., Yoshida, H., and Kobayashi, K. 2014. Development of a Stationary Robotic Strawberry Harvester with a Picking Mechanism that Approaches the Target Fruit from Below. Japan Agricultural Research Quarterly: JARQ, 48(3), pp.261-269.
\bibitem{c2} Dimeas, F., Sako, D.V., Moulianitis, V.C. and Aspragathos, N.A., 2015. Design and fuzzy control of a robotic gripper for efficient strawberry harvesting. Robotica, 33(5), pp.1085-1098.
\bibitem{c3} Hayashi, S., Shigematsu, K., Yamamoto, S., Kobayashi, K., Kohno, Y., Kamata, J. and Kurita, M., 2010. Evaluation of a strawberry-harvesting robot in a field test. Biosystems engineering, 105(2), pp.160-171.
\bibitem{c4} Van Henten, E.J., Van’t Slot, D.A., Hol, C.W.J. and Van Willigenburg, L.G., 2009. Optimal manipulator design for a cucumber harvesting robot. Computers and electronics in agriculture, 65(2), pp.247-257.
\bibitem{c5} Eizicovits, D., van Tuijl, B., Berman, S. and Edan, Y., 2016. Integration of perception capabilities in gripper design using graspability maps. Biosystems Engineering, 146, pp.98-113.
\bibitem{c6} Chiu, Y.C., Yang, P.Y. and Chen, S., 2013. Development of the end-effector of a picking robot for greenhouse-grown tomatoes. Applied engineering in agriculture, 29(6), pp.1001-1009.
\bibitem{c7} Lehnert, C., English, A., McCool, C., Tow, A.W. and Perez, T., 2017. Autonomous sweet pepper harvesting for protected cropping systems. IEEE Robotics and Automation Letters, 2(2), pp.872-879.
\bibitem{c8} Hayashi, S., Takahashi, K., Yamamoto, S., Saito, S. and Komeda, T., 2011. Gentle handling of strawberries using a suction device. Biosystems engineering, 109(4), pp.348-356.
\bibitem{c9} Sa, I., Lehnert, C., English, A., McCool, C., Dayoub, F., Upcroft, B. and Perez, T., 2017. Peduncle detection of sweet pepper for autonomous crop harvesting—Combined Color and 3-D Information. IEEE Robotics and Automation Letters, 2(2), pp.765-772.
\bibitem{c10} Huang, Z., Wane, S. and Parsons, S., 2017, July. Towards Automated Strawberry Harvesting: Identifying the Picking Point. In Conference Towards Autonomous Robotic Systems(pp. 222-236). Springer, Cham.
\bibitem{c11} Ficken, L.A., Unidynamics Corporation, 1989. Iris mechanism. U.S. Patent 4,804,108.
\bibitem{c12} Takaki, T. and Omata, T., 2011. High-performance anthropomorphic robot hand with grasping-force-magnification mechanism. IEEE/ASME Transactions on Mechatronics, 16(3), pp.583-591.
\bibitem{c13} Dong, X., Raffles, M., Guzman, S.C., Axinte, D. and Kell, J., 2014. Design and analysis of a family of snake arm robots connected by compliant joints. Mechanism and Machine Theory, 77, pp.73-91.
\bibitem{c14} Grimstad, L. and From, P.J., 2017. The Thorvald II agricultural robotic system. Robotics, 6(4), p.24.
\bibitem{c15} Grimstad, L., Pham, C.D., Phan, H.T. and From, P.J., 2015, June. On the design of a low-cost, light-weight, and highly versatile agricultural robot. In Advanced Robotics and its Social Impacts (ARSO), 2015 IEEE International Workshop on (pp. 1-6). IEEE.
\bibitem{c16} Bac, C.W., Hemming, J., Tuijl, B.A.J., Barth, R., Wais, E. and Henten, E.J., 2017. Performance Evaluation of a Harvesting Robot for Sweet Pepper. Journal of Field Robotics.



\end{thebibliography}
\end{document}